\PassOptionsToPackage{table,dvipsnames}{xcolor}
\documentclass[10pt,twocolumn,letterpaper]{article}

\usepackage[pagenumbers]{cvpr} 

\definecolor{cvprblue}{rgb}{0.21,0.49,0.74}
\usepackage[pagebackref,breaklinks,colorlinks,allcolors=cvprblue]{hyperref}
\usepackage{pifont}
\usepackage{placeins}
\usepackage{pgfplots}
\pgfplotsset{compat=1.18}

\newcommand{\xmark}{\textcolor{red}{\ding{55}}} 
\newcommand{\cmark}{\textcolor{green}{\ding{51}}} 
\usepackage{multirow}
\usepackage{pgfplots}
\usepackage{graphicx}
\usepackage{bm} 
\usepackage{footmisc}

\def\paperID{****} 
\def\confName{CVPR}
\def\confYear{2026}

\title{GHOST: Fast Category-agnostic Hand-Object Interaction Reconstruction from RGB Videos using Gaussian Splatting}

\author{
Ahmed Tawfik Aboukhadra$^{1,2}$ \and
Marcel Rogge$^{1,2}$ \and
Nadia Robertini$^{2}$ \and
Abdalla Arafa$^{1,2}$ \and
Jameel Malik$^{4}$  \and
Ahmed Elhayek$^{3}$ \and
Didier Stricker$^{1,2}$ \and \\
{ $^{1}$RPTU$\;\;$
$^{2}$DFKI-AV Kaiserslautern$\;\;$
$^{3}$UPM Saudi Arabia$\;\;$  
$^{4}$NUST-SEECS Pakistan$\;\;$} \\
{\tt \small firstname\{\_secondname\}.lastname@dfki.de}
\vspace{-5mm}
}

\begin{document}

\maketitle
\begin{abstract}

Understanding realistic hand–object interactions from monocular RGB videos is essential for AR/VR, robotics, and embodied AI. 
Existing methods rely on category-specific templates or heavy computation, yet still produce physically inconsistent hand–object alignment in 3D.
We introduce $\textbf{GHOST}\ (\textit{\textbf{G}aussian \textbf{H}and-\textbf{O}bject \textbf{S}pla\textbf{T}ting})$, a fast, category-agnostic framework for reconstructing dynamic hand-object interactions using $\textit{2D Gaussian Splatting}$. 
GHOST represents both hands and objects as dense, view-consistent Gaussian discs and introduces three key innovations: 
(1) a geometric-prior retrieval and consistency loss that completes occluded object regions, 
(2) a grasp-aware alignment that refines hand translations and object scale to ensure realistic contact, and 
(3) a hand-aware background loss that prevents penalizing hand-occluded object regions. 
GHOST achieves complete, physically consistent, and animatable reconstructions from a single RGB video while running an order of magnitude faster than prior category-agnostic methods. 
Extensive experiments on ARCTIC, HO3D, and in-the-wild datasets demonstrate state-of-the-art accuracy in 3D reconstruction and 2D rendering quality, establishing GHOST as an efficient and robust solution for realistic hand-object interaction modeling. 
Code is available at \url{https://github.com/ATAboukhadra/GHOST}.
\vspace{-3mm}

\end{abstract}

\begin{figure}
    \centering
    \includegraphics[width=\linewidth]{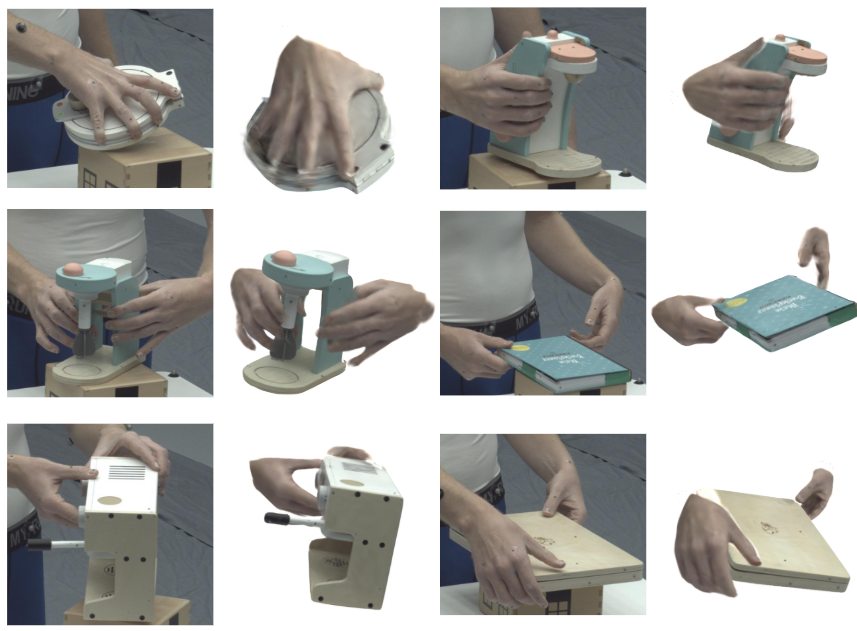}
    \caption{
Our method reconstructs complete 3D hand–object interactions from a single monocular RGB video—recovering full object surfaces and realistic hand contact even under severe occlusions—while enabling fast, accurate, category-agnostic reconstruction and novel-view rendering.
\vspace{-3mm}
    }
    \label{fig:teaser}
\end{figure}

\section{Introduction}

Hands enable nearly every form of physical interaction we perform, such as grasping, manipulating objects, operating devices, or expressing intent~\cite{han2025touch,huang2025hoigpt,aboukhadra2024surgeonet,YangCVPR2022OakInk,zhan2024oakink2}. 
As a result, reconstructing realistic 3D hand–object interactions from monocular RGB videos has become fundamental for virtual reality (VR), augmented reality (AR), teleoperation, and embodied AI. 
Achieving this, however, remains challenging: mutual occlusions between hands and objects, large variation in object topologies, and inherent depth ambiguities in monocular settings often lead to unstable scales and physically inconsistent contact.

Although significant progress has been made, existing hand-object reconstruction methods typically fall into two categories. 
Template-based approaches~\cite{hasson2020leveraging,hasson2021towards,chen2023gsdf} can generate high-quality meshes, but are restricted to a small, fixed set of known object shapes, restricting their ability to generalize to novel or unconventional objects encountered in real-world scenarios. 
Category-agnostic methods remove this constraint but are often computationally expensive~\cite{fan2024hold,on2025bigs}, struggle to recover missing geometry under occlusion~\cite{aboukhadra2023thornet}, or require multi-view supervision~\cite{manus24}. 
NeRF-based pipelines such as HOLD \cite{fan2024hold} achieve photorealistic rendering, yet demand hours of per-sequence optimization. 
Recent Gaussian-splatting frameworks improve runtime and provide explicit 3D representations \cite{on2025bigs,manus24}. 
However, these methods still tend to produce unrealistic hand–object contact under severe occlusions, and their optimization remains time-consuming, limiting practical applicability.

To address these challenges, we present \textbf{GHOST}, a fast, category-agnostic framework for reconstructing realistic bimanual hand-object interactions from monocular RGB videos. 
GHOST represents hands and objects as dense, view-consistent Gaussian discs, enabling complete and physically consistent reconstructions even under heavy occlusions. 
By integrating grasp-aware reasoning with geometric object priors to recover hidden surfaces, GHOST achieves high-quality, photorealistic rendering and produces animatable reconstructions. 
The framework operates in three stages: 
(1) Preprocessing, which initializes hand meshes and retrieves object priors; 
(2) hand-object alignment, which refines scale and hands translation through grasp-aware reasoning; and 
(3) Gaussian-splatting optimization, which jointly reconstructs hands and objects with occlusion-aware consistency. 
Together, these components enable GHOST to achieve state-of-the-art performance in both 3D accuracy and 2D rendering fidelity (See Fig.~\ref{fig:teaser}) while running over an order of magnitude faster than previous category-agnostic methods.
%
%
Our contributions are summarized as follows:
\begin{itemize}
\item A fast, category-agnostic framework for reconstructing animatable bimanual hand-object interactions from monocular RGB sequences.
\item A prior-aware reconstruction strategy that fills occluded object regions using geometric consistency from retrieved priors.
\item A grasp-aware alignment that ensures realistic and stable hand-object contact.
\item A hand-aware background loss that preserves valid object regions despite persistent occlusion.
\item State-of-the-art performance on both 3D reconstruction and 2D rendering metrics, while achieving over 13× faster runtime than prior category-agnostic approaches.
\end{itemize}
We conduct an extensive evaluation across three datasets: ARCTIC, HO3D, and in-the-wild videos, covering both 3D geometric and 2D photometric metrics. 
GHOST consistently surpasses previous methods in hand-object reconstruction accuracy and rendering quality, establishing a new benchmark for fast and realistic category-agnostic hand-object reconstruction.

\section{Related Work}

The problem of realistic hand–object reconstruction spans multiple subdomains, including monocular hand reconstruction under occlusion, animatable avatars, and category-agnostic interaction reconstruction.

\paragraph{3D Hand Reconstruction from RGB under Occlusions.}
Monocular hand reconstruction has been extensively studied in~\cite{pavlakos2024reconstructing,potamias2025wilor,dong2024hamba,aboukhadra2023thornet}. 
Recent Transformer-based approaches~\cite{park2022handoccnet,lin2021endtoend,pavlakos2024reconstructing,potamias2025wilor,aboukhadra2023thornet} significantly improve robustness under occlusions, achieving accurate 3D hand meshes. 
HaMeR~\cite{pavlakos2024reconstructing}, in particular, provides a strong baseline trained on a large-scale dataset and generalizes well to in-the-wild scenarios, making it suitable for initializing hands in interaction pipelines.

\paragraph{Neural and Gaussian-based Avatars.}
Neural implicit representations such as NeRFs~\cite{mildenhall2020nerf}, and the more recent explicit formulations like Gaussian Splatting~\cite{kerbl2023gaussiansplatting}, have become standard for photorealistic, animatable human avatars~\cite{chen2023hand,qian2024gaussianavatars,Pang_2024_CVPR}. 
Recent works~\cite{moreau2024human,kocabas2024hugs,Pang_2024_CVPR,shao2024splattingavatar} initialize a set of 3D Gaussians on the surface of a canonical SMPL~\cite{loper2023smpl} or FLAME~\cite{li2017learning} mesh and animate them through parameter-controlled skeletal motion. 
Rendering is performed by projecting the Gaussians onto the image plane, where they are blended to produce smooth, photorealistic images of the avatar.


\paragraph{Hand-Object Interaction Reconstruction.}
Objects, unlike hands, lack consistent topology or parametric structure and exhibit wide variations in shape, material, and appearance. 
Early approaches~\cite{hasson2019learning,hasson2020leveraging,hasson2021towards} jointly reconstructed hand and object meshes from RGB inputs, enforcing contact through attraction–repulsion losses and leveraging photometric or silhouette consistency. 
However, these methods rely on fixed object templates, limiting generalization. 
Template-free representations such as THOR-Net~\cite{aboukhadra2023thornet} and ShapeGraformer~\cite{aboukhadra2024shapegraformer} addressed this limitation by modeling objects as spherical deformations with shared topology, enabling category-agnostic learning.

\paragraph{Category-agnostic Hand-Object Reconstruction.}
Building on implicit neural fields~\cite{mildenhall2020nerf}, several category-agnostic pipelines~\cite{fan2024hold,ye2022s,ye2023diffusion} reconstruct hand–object interactions directly from monocular sequences. 
HOLD~\cite{fan2024hold} uses off-the-shelf hand pose estimators~\cite{lin2021endtoend,pavlakos2024reconstructing}, SAM-based object segmentation~\cite{kirillov2023segment,cheng2023segment}, and HLoc-based SfM~\cite{sarlin2020superglue,schoenberger2016sfm,sarlin2019coarse} to recover geometry via NeRF optimization, but requires hours per sequence. 
BIGS~\cite{on2025bigs} replaces the NeRF stage with Gaussian Splatting~\cite{kerbl2023gaussiansplatting}, slightly improving runtime and explicitness of the representation. 
However, these methods still produce unrealistic contact under occlusions and high computational cost, motivating our efficient and physically consistent Gaussian-based framework.

\begin{figure*}[t]
    \centering

    \includegraphics[width=\textwidth]{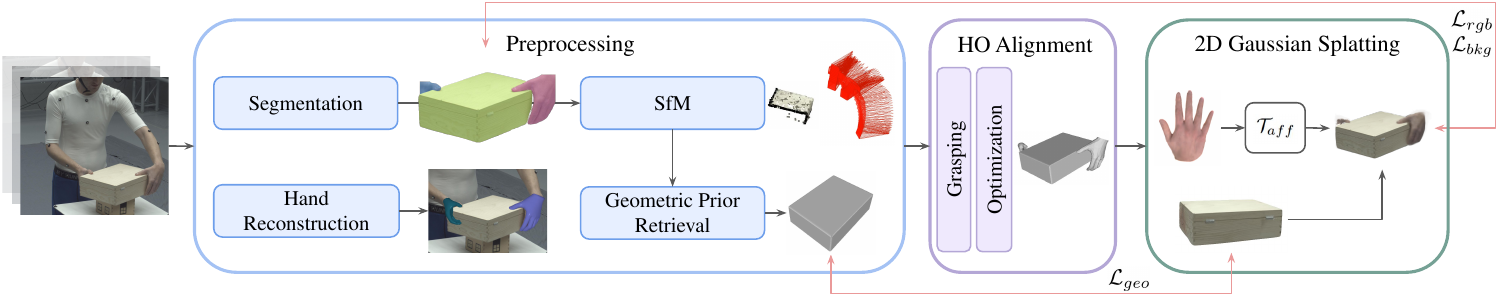}

    \hspace{1.7cm}
    \includegraphics[width=0.85\textwidth]{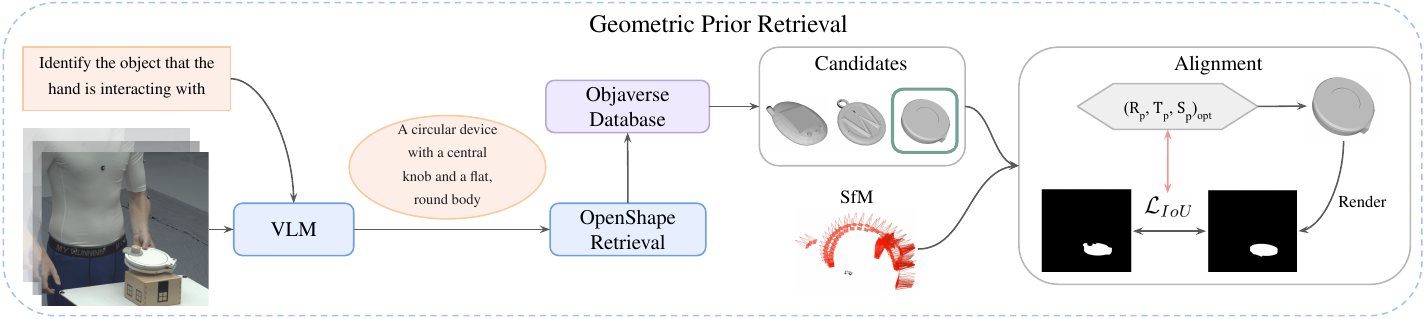}

    \caption{
    (Top) Overview of our pipeline, which consists of three stages. 
    In \textbf{preprocessing}, we extract hand meshes, camera poses, and object information (i.e. mask, point cloud, and geometric prior). 
    During \textbf{hand–object alignment}, object's scale and hand translations are optimized using grasp-aware and temporal reasoning. 
    In the \textbf{Gaussian Splatting stage}, hands and objects are jointly reconstructed with occlusion-aware losses. 
    (Bottom) Retrieval and alignment pipeline used to obtain the object’s geometric prior.
    }
    \label{fig:pipeline}
\end{figure*}


    



\section{Method}
Our method reconstructs hand–object interactions from monocular RGB videos. 
It consists of three main stages: a preprocessing pipeline that extracts geometric and motion cues, a HO alignment stage that uses tracking and grasp-aware reasoning to align hands with the object, and a Gaussian Splatting optimization stage that jointly reconstructs the photorealistic hands and object in 3D. 
Fig.~\ref{fig:pipeline} summarizes the stages of our pipeline. 
In the following subsections, we describe each component in detail.


\subsection{Preprocessing Pipeline}


Given an input video ${V} = \{ I_t \}_{t=1}^T$, with $T$ frames, we first segment the object using SAM2~\cite{ravi2024sam2}, obtaining per-frame masks $\mathcal{M}_t^{o}$. 
Using the object's segmentation, we apply Structure-from-Motion (SfM) to estimate camera intrinsics $\mathbf{K}$, relative camera poses $\{(\mathbf{R}_t^c, \mathbf{T}_t^c)\}_{t=1}^T$, and a sparse object point cloud $\mathcal{P}_{sfm}$. 
For each frame $t$, the full camera projection function is denoted as: 
$\Pi_t(\mathbf{x}) = \mathbf{K}[\mathbf{R}_t^c \mid \mathbf{T}_t^c][\mathbf{x}]$
Given the critical influence of SfM on subsequent stages, we compare two SfM approaches. 
In the first, following HOLD~\cite{fan2024hold}, we employ HLoc~\cite{sarlin2019coarse} with COLMAP~\cite{schoenberger2016sfm,schoenberger2016mvs}, enhanced by a temporal-window pairing strategy to improve candidate matching. 
In parallel, we evaluate the recent VGGSfM~\cite{wang2024vggsfm} video-based pipeline and demonstrate its benefits for the ARCTIC~\cite{fan2023arctic} dataset.

\subsubsection{Object Geometric Prior}
Reconstructing objects under hand manipulation is challenging due to frequent hand and self-occlusions, which often leave unseen regions incomplete. 
To address this, we leverage open-source large-scale 3D object databases such as Objaverse~\cite{deitke2023objaverse,deitke2023objaversexl}, which provide extensive coverage of everyday objects. 
These models serve as geometric priors to refine and complete partial reconstructions, thereby improving both object accuracy and hand-object interaction quality. 
Fig.~\ref{fig:pipeline} (Bottom part) shows our geometric prior retrieval and alignment algorithm.

\paragraph{Retrieval.}
A textual description $d$ of the hand-held object is obtained from a few sampled video frames using a vision-language model (e.g., InternVL~\cite{chen2024internvl,chen2024far}) or provided directly (e.g., 'Box'). 
The text embedding $\phi_{txt}(d)$ is computed using OpenShape's~\cite{liu2023openshape} CLIP model, which maps text and object meshes into a shared embedding space. 
Since all Objaverse objects are pre-embedded, retrieval reduces to a nearest-neighbor search between $\phi_{txt}(d)$ and stored embeddings $\{\phi_{obj}(\mathcal{O}_p)\}$, producing $k$ candidate meshes $\{\mathcal{O}_1, \dots, \mathcal{O}_k\}$. 
We use a ray casting algorithm to simplify these structures and extract their outer surface.

\paragraph{Prior-Mask Alignment.}
The retrieved 3D meshes are not aligned with the object’s point cloud $\mathcal{P}_{sfm}$; they may be rotated, shifted, or scaled differently. 
Therefore, we estimate an affine transformation that includes quaternion rotation $\mathbf{R}_{p}$, 3D translation $\mathbf{T}_{p}$, and 3D scale $\mathbf{S}_{p}$ for each retrieved candidate mesh $\mathcal{O}_p$.
We optimize previous variables such that, after transforming the mesh and projecting it into the camera view, the rendered mesh silhouette matches the object's mask $\mathcal{M}_t^o$. 
This alignment is measured using the Intersection-over-Union loss $\mathcal{L}_{IoU}$ between the rendered silhouette and the ground-truth mask. 
More formally:
\begin{equation}    
\mathcal{L}_{IoU} = 1 - \text{IoU}\big(\Pi_t([\mathbf{R}_{p}|\mathbf{T}_{p}] [\mathbf{S}_{p}\cdot\mathcal{O}_p]), \mathcal{M}_t^{o}\big).
\end{equation}
For all subsequent optimization stages, we select the object's geometric prior $\mathcal{O} \in \mathbb{R}^{N \times 3}$ with $N$ vertices as the transformed candidate $[\mathbf{R}_{p}|\mathbf{T}_{p}][\mathbf{S}_{p}\cdot\mathcal{O}_p]$ with the lowest $\mathcal{L}_{IoU}$ . 
If the retrieved prior has a low number of vertices, we densify its point cloud by adding more points on its surface.

\subsubsection{Hand Reconstruction Initialization}
\label{sec:hamer}
We obtain initial 3D meshes of both hands using an off-the-shelf hand reconstruction model (i.e., HaMeR~\cite{pavlakos2024reconstructing}), guided by bounding-box detections from RTMPose~\cite{jiang2023rtmpose,jiang2023rtmlib,jiang2024rtmw}. 
For each frame $I_t$, HaMeR produces MANO~\cite{romero2017mano} parameters consisting of pose $\theta_t$, shape $\beta_t$, and global hand rotation $\mathbf{R}_t^{h}$. 
In practice, strong occlusions from the manipulated object often result in jittery or implausible hand reconstructions. 
To address this issue, we apply a post-processing step that detects and removes unreliable frames. 
Frames whose MANO parameters deviate beyond predefined thresholds from their neighboring frames are discarded, and their parameters are subsequently recomputed using linear interpolation for translations and spherical interpolation for rotations\footnote{See the supplementary for more details.\label{supp}}.
This process produces temporally consistent 3D hand vertices $\mathcal{V}_t^{h}$, 3D joints $\mathcal{J}_{t}^{h}$, and 2D joints $\mathcal{J}_{\text{2D, }t}^{h}$ for left and right hands $h \in \{r,l\}$ throughout the entire sequence. 
%
In addition, we segment and track the hand masks using SAM2~\cite{ravi2024sam2} and crop it using the hand's bounding boxes to exclude the forearm, producing $\mathcal{M}_t^{h}$.



\subsection{Hand Translation and Object Scale Optimization (HO Alignment)}

Initial hand reconstructions and the object's point cloud $\mathcal{P}_{sfm}$ may be misaligned in 3D, due to scale inconsistencies from SfM or drifting hand translations estimated by HaMeR. 
To correct this, we jointly refine global hand translations $\mathbf{T}_t^{h}$ and global object scale $\mathbf{S}_o$. 
We describe the details of this step in the following subsections. 
The optimized scale is then applied to the point cloud $\mathcal{P}_{sfm}$ and camera translations $\mathbf{T}_t^c$.

%

\subsubsection{Grasping Detection}
In bimanual hand-object interactions, changes to object poses are the result of hand interaction, and hence, the hand and object should be in close contact when the object is moving. 
To identify frames where the hands interact with the object, we compare the motion trajectories of both hands relative to the object's motion. 
The object's motion $\Delta \mathcal{T}^{o}$ is computed as the change in its point cloud center across frames.
For each hand, we compute its translation change $\Delta \mathcal{T}^{h}$ and measure the cosine similarity between its motion and the object's motion, projected to the $x,y$ plane:
\begin{equation}
c_{h} = \, \Delta \hat{\mathcal{T}}^{o}_{xy} \cdot \Delta \hat{\mathcal{T}}^{h}_{xy},
\end{equation}
where $\Delta \hat{\mathcal{T}}$ denotes normalized translation direction vectors. 
If both hands have $c_{h} > \tau_{sim}$, the frame is labeled as a two-hand grasp; otherwise, the hand with the higher score is considered the interacting one. 
We empirically choose $\tau_{sim}=0.5$.

\subsubsection{Optimization Objective}
Once grasping frames are identified, we optimize a joint objective consisting of three complementary loss terms.

\paragraph{Contact Loss ($\mathcal{L}_{contact}$).}
Applied only during grasping frames, the contact loss enforces the interacting hand to remain close to the object. 
It is defined as the Chamfer Distance (CD) between the translated hand mesh vertices $\mathcal{V}_t^{h}$ and the scaled object geometric prior $\mathcal{O}$. 
If only one hand is grasping, the loss is applied to that hand’s vertices. 
If both hands grasp, both sets of vertices are included.

\begin{equation}    
\mathcal{L}_{contact} = {\text{CD}}(\mathbf{T}_t^h + \mathcal{V}_t^{h}, [\mathbf{R}_t^c \mid \mathbf{T}_t^c][\mathbf{S}_o \cdot \mathcal{O}]).
\end{equation}


\paragraph{Hand 2D Projection Loss ($\mathcal{L}_{proj}$).}
This loss aims to use the camera intrinsics $\mathbf{K}$ to compute reprojection error between translated hand joints $\mathcal{J}_t^h$ and the HaMeR-predicted 2D hand joints and it is defined as:
\begin{equation}
\mathcal{L}_{\text{proj}} = 
\left\| 
\Pi_t \left( \mathbf{T}_t^h + \mathcal{J}_t^h \right)
- 
\mathcal{J}_{\text{2D, }t}^h
\right\|_1.
\end{equation}
In practice, small bounding boxes typically indicate hands that are far from the camera and thus less reliable. 
Such low-confidence predictions are omitted when computing $\mathcal{L}_{\text{proj}}$. 
The purpose of this term is to maintain 2D consistency of the hands.

\paragraph{Temporal Consistency Loss ($\mathcal{L}_{temp}$).}
Enforces smooth hand translation across time defined as:
\begin{equation}
\mathcal{L}_{\text{temp}} = 
\frac{1}{T-1} \sum_{t=1}^{T-1} 
\left\| \mathbf{T}_{t+1}^h - \mathbf{T}_{t}^h \right\|_2^2.
\end{equation}
Combined together, they form:
\begin{equation}
   \mathcal{L} = \lambda_1 \mathcal{L}_{contact} + 
   \lambda_2 \mathcal{L}_{proj} +
   \lambda_3 \mathcal{L}_{temp}.
\end{equation}
We empirically choose $\lambda_1=10^{3}, \lambda_2=10^{-1}, \text{ and } \lambda_3=10$ to balance the loss term values\footnotemark[1].

\subsection{Hand-Object Gaussian Splatting}

\subsubsection{Preliminary}

Gaussian Splatting (GS)~\cite{kerbl2023gaussiansplatting} represents static 3D scenes from calibrated multi-view images and enables rendering from novel viewpoints. 
Instead of optimizing a neural volume, the scene is modeled as a set of 3D Gaussians, each defined by its center, rotation, scale, opacity, and color encoded in spherical harmonics. 
During optimization, a differentiable rasterizer renders RGB views that are compared against ground-truth views using a photometric loss term consisting of an L1 and D-SSIM term, denoted in Fig.~\ref{fig:pipeline} as $\mathcal{L}_{rgb}$. 
The number of Gaussians adaptively changes through \emph{densification} (adding Gaussians in sparse regions) and \emph{pruning} (removing excessively large or low-opacity ones).

\subsubsection{Object Optimization}

Rogge~\etal~\cite{RoggeOC2DGS2025} applied 2D Gaussian Splatting~\cite{huang20242d} to reconstruct segmented objects from RGB videos. 
To suppress Gaussians rendered outside object masks, they applied a background loss $\mathcal{L}_{bkg}$ that penalizes background-projected Gaussians during training. 
Following their setup, we fit Gaussians to the object in the first stage of our pipeline, where the set of all Gaussians is parameterized as,
\begin{equation}
\mathcal{G}_o = \{ c_o, r_o, s_o, \alpha_o, SH_o \}.
\end{equation}
Here, $c_o$ are 3D centers, $r_o$ are 3D rotations, $s_o$ are 2D scales, $\alpha_o$ are opacities, and $SH_o$ are spherical harmonics that represent the color of the Gaussians. 
We further introduce two losses tailored for hand–object interaction. 
We explain these losses and their rationale in the following paragraphs.

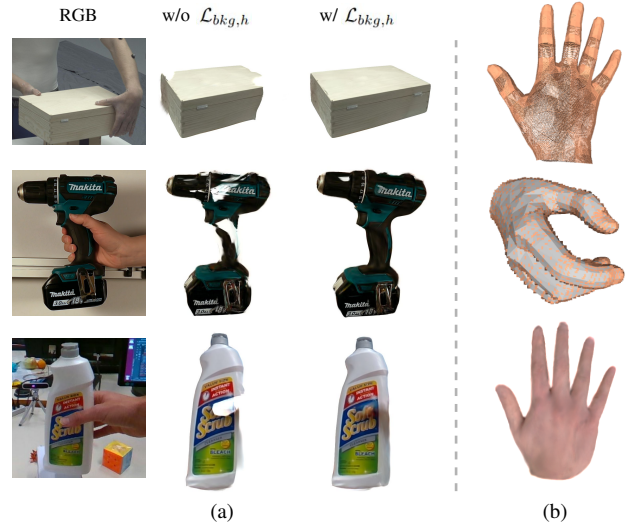
\begin{figure}[t]
    \centering
    \begin{subfigure}[t]{0.68\columnwidth}
        \centering
        \includegraphics[width=\linewidth]{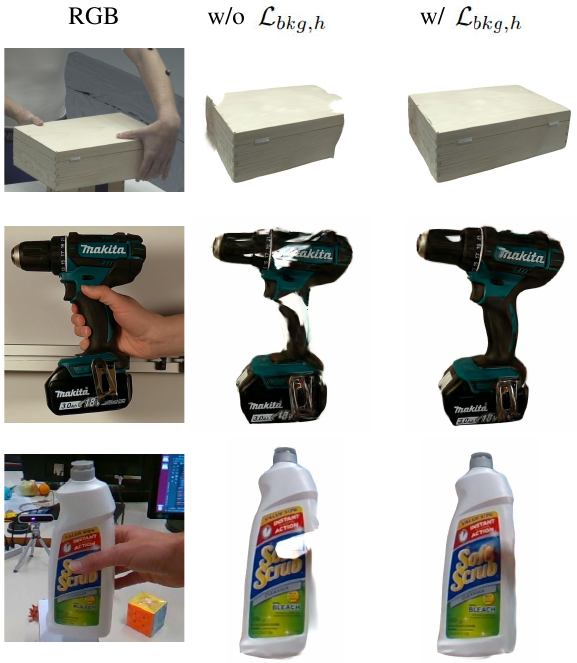}
        \caption{}
        \label{fig:ablation_bkg}
    \end{subfigure}
    \hspace{0.01\columnwidth}
    \begin{tikzpicture}
        \draw[dashed, gray!60, line width=1.0pt] (0,0) -- (0,6);
    \end{tikzpicture}
    \hspace{0.01\columnwidth}
    \begin{subfigure}[t]{0.25\columnwidth}
        \centering
        \includegraphics[width=\linewidth]{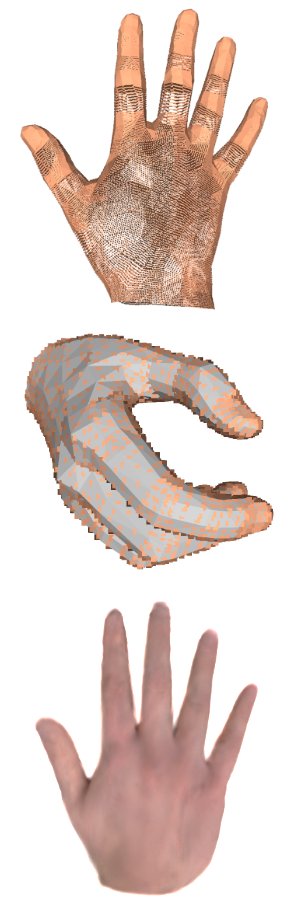}
        \caption{}
        \label{fig:hand_gaussians}
    \end{subfigure}

    \caption{
       Qualitative comparison showing (a)  the effect of our novel background loss $\mathcal{L}_{bkg,h}$ on object reconstruction quality. 
        (b) \textbf{Top}: Gaussian centers on the canonical hand; \textbf{middle}: deformed hand mesh with aligned Gaussian centers after $\mathcal{T}_{aff}$; \textbf{bottom}: final animatable Gaussian hand after training.
    }
    \label{fig:combined_ablation_hand}
\end{figure}

\paragraph{Hand-aware Background Loss ($\mathcal{L}_{bkg,h}$).}
While effective in single-object scenes, the background loss $\mathcal{L}_{bkg}$ is less suited for cases with strong occlusions, such as when a hand holds the object. 
In such cases, occluded regions may be incorrectly removed (see finger-shaped holes in Fig.~\ref{fig:ablation_bkg}). 
To mitigate this, we combine both hands and object masks to more reliably distinguish foreground from background during object reconstruction creating an updated hand-aware background loss term ($\mathcal{L}_{bkg,h}$). 
Using hand masks ensures that occluded object regions are retained rather than mistakenly discarded as background. 
Although hand masks inevitably include some non-object areas, the temporal variation across frames -where the object and hands appear under different orientations- guarantees that all true background regions are eventually removed over time.

\paragraph{Geometric Consistency Loss ($\mathcal{L}_{geo}$).}
Using the geometric prior $\mathcal{O}$ from earlier stages, we introduce a novel geometric consistency loss $\mathcal{L}_{geo}$ to keep the reconstructed Gaussians consistent with the prior surface. 
The first component, $\mathcal{L}_{out}$, penalizes Gaussians whose centers are farther than a threshold $\tau_{out}$ from the prior, preventing them from drifting away from the object surface. 
The second component, $\mathcal{L}_{fill}$, measures the distance from each point on the prior surface to the closest Gaussian center and penalizes large gaps that exceed a threshold $\tau_{fill}$, encouraging the model to fill holes caused by occlusions. 
The final loss is a weighted combination where:
\begin{equation}
\mathcal{L}_{\text{out}} = \sigma\big(\|{c}_{o} - \mathcal{O}\|_2 - \tau_{\text{out}}\big)^2,
\mathcal{L}_{\text{fill}} = \sigma\big(\|\mathcal{O} - {c}_{o}\|_2 - \tau_{\text{fill}}\big)^2,
\end{equation}
and
\begin{equation}
\mathcal{L}_{geo} = \lambda_{geo} (\mathcal{L}_{out} + \mathcal{L}_{fill}),
\end{equation}
where $\sigma$ is the ReLU operation and $\lambda_{geo}$ is empirically set to $5$. 
In essence, $\mathcal{L}_{geo}$ keeps Gaussians close to the prior while promoting completeness in the reconstructed surface.

\subsubsection{Hand Optimization}
\label{sec:rigging}
Unlike rigid objects, hands deform over time, making Gaussian Splatting more challenging. 
To handle this, we follow GaussiansAvatars~\cite{qian2024gaussianavatars} and represent hands in a canonical space by attaching Gaussians to the MANO~\cite{romero2017mano} mesh faces and, similar to the object, represent canonical hand Gaussians as:
\begin{equation}
\mathcal{G}_h = \{ c_h, r_h, s_h, \alpha_h, SH_h \}.
\end{equation}

\begin{figure*}[t]
    \centering
    \includegraphics[width=0.9\textwidth]{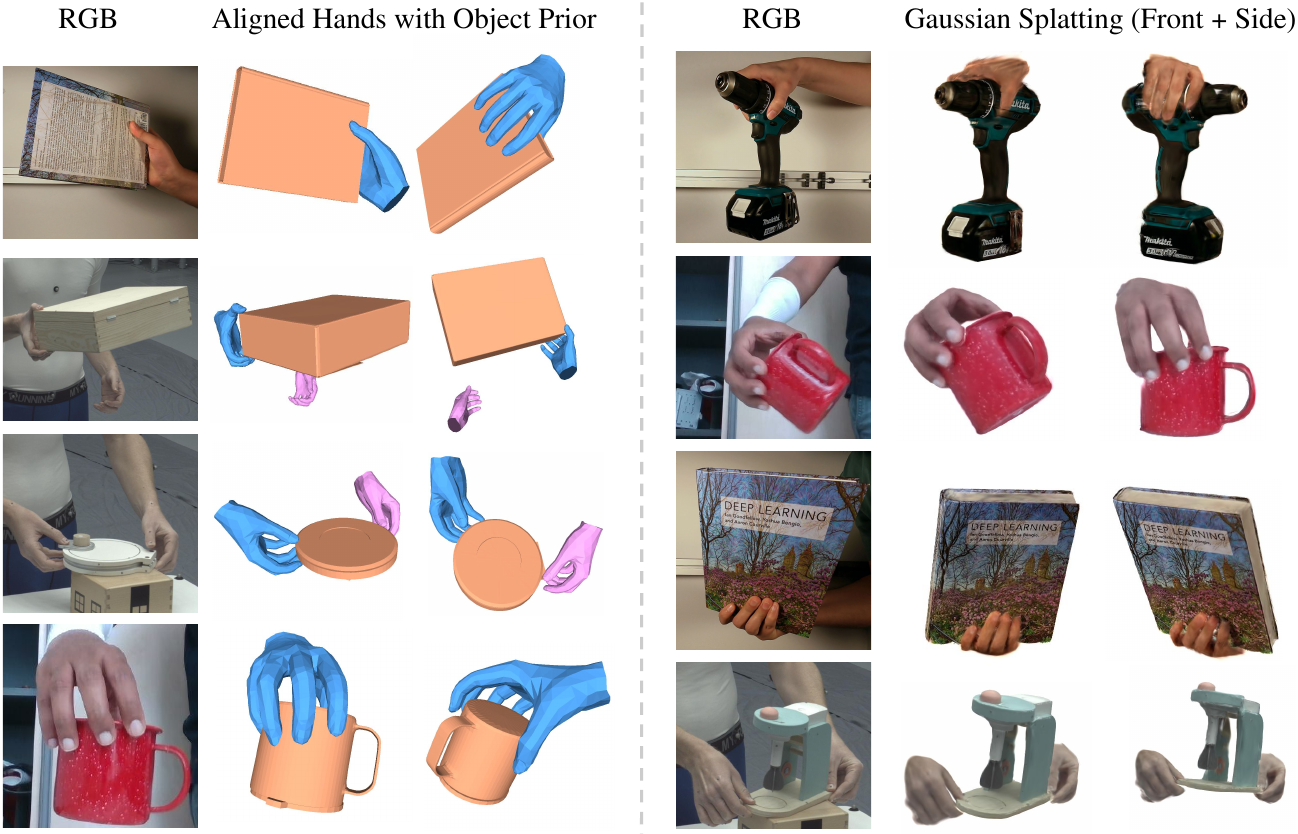}
    \caption{
    Qualitative results of GHOST on ARCTIC~\cite{fan2023arctic}, HO3D~\cite{hampali2020honnotate}, and in-the-wild examples: 
    \textbf{Left}: shows aligned 3D hand meshes with the object’s geometric prior obtained during HO alignment.
    \textbf{Right:} presents photorealistic Gaussian Splatting renderings from original viewpoint and novel viewpoints.
    GHOST produces consistent hand–object alignment in 3D and maintains realistic appearance even under view changes, enabling physically plausible interaction reconstruction and high-fidelity rendering across viewpoints. 
    }
    \label{fig:qualitative}
\end{figure*}

\paragraph{Hand Gaussians Rigging.}
Let the canonical MANO mesh be defined by vertices 
$\mathcal{V} \in \mathbb{R}^{778 \times 3}$ 
and faces 
$\mathcal{F} \in \mathbb{R}^{1538 \times 3}$. 
For each face $f \in \mathcal{F}$, we attach $m$ Gaussians with 
canonical centers $c_{h,f} \in \mathbb{R}^{\text{m} \times 3}$, 
rotations $r_{h,f} \in SO(3)^m$, 
and scales $s_{h,f} \in \mathbb{R}^{m \times 2}$. 
Given deformed hand meshes $\mathcal{V}^h_t$, we compute for each face a local affine transformation that maps from its canonical to deformed state. 
This transformation $\mathbf{M}_{f,t}$ is derived from basis matrices constructed over the face vertices in the canonical pose ($\mathbf{B}_f^c$) and the deformed pose ($\mathbf{B}_{f,t}^d$):
\begin{equation}
\mathbf{M}_{f,t} = \mathbf{B}_{f,t}^d \, \mathbf{B}_f^{c^{-1}}.
\end{equation}
Each Gaussian $i$ attached to face $f$ then deforms according to the following equations:
\begin{equation}
c_{h,f,i}^d = \mathbf{M}_{f,t} \, c_{h,f,i}, 
\quad 
r_{h,f,i}^d = \text{Polar}\!\left(\mathbf{M}_{f,t}\right),
\end{equation}
where $\text{Polar}(\cdot)$ extracts the rotation from $\mathbf{M}_{f,t}$. 
This formulation allows Gaussians to follow the mesh deformation of their corresponding faces, as shown in Fig.~\ref{fig:hand_gaussians}. 
We refer to this operation by $\mathcal{T}_{aff}$.

Finally, we optimize the hand Gaussians by loading the pretrained object representation $\mathcal{G}_o$ into the 3D scene and optimizing only $(s_h, \alpha_h, SH_h)$, while keeping mesh-driven parameters ($c_h$ and $r_h$) fixed. 
We also allow the hand translation ($\mathbf{T}_t^h$) to change during optimization to create more 2D consistency for the rendered hand. 
We disable the steps of pruning and densification in the joint hand-object optimization and study the effect of $m$ on the rendering quality. 
This two-stage optimization—object first, then deformable hands—produces a coherent joint representation of hands and object renderable from arbitrary viewpoints.

\section{Experiments and Results}
In this section, we describe our experimental setup, including the datasets, evaluation metrics, ablation studies, and comparisons with state-of-the-art methods.

\subsection{Datasets}

We evaluate our method on three sources of data. 
First, we use the \textbf{ARCTIC Bi-CAIR} dataset~\cite{fan2023arctic,fan2024hold}, which provides $9$ video sequences of two hands interacting with diverse objects. 
Second, we evaluate sequences from the \textbf{HO3D} dataset~\cite{hampali2020honnotate}. 
This benchmark provides a controlled setup for assessing category-agnostic object reconstruction under severe hand occlusions. 
Finally, we capture two additional hand–object interaction sequences (Drill and Book) using a GoPro camera.
These videos are used for qualitative evaluation to demonstrate the generalization of our method beyond existing datasets.

\subsection{Metrics}

All 3D evaluation metrics follow the official protocol released by HOLD~\cite{fan2024hold} for the ARCTIC~\cite{fan2023arctic} Bi-CAIR challenge\footnote{\url{https://arctic-leaderboard.is.tuebingen.mpg.de/leaderboard}}. 
For hands, we report the mean per-joint position error (MPJPE), computed on root-aligned hand joints. 
For objects, we evaluate multiple Chamfer-based metrics. 
Object's Chamfer Distance relative to each hand’s root (CD$_r$, CD$_l$) quantify the interaction quality between the hand and object surfaces, while also being sensitive to the object’s absolute scale, as no rigid alignment is applied. 
Their average forms CD$_h$. 
We further compute CD$_{ICP}$, which aligns predicted and ground-truth object point clouds via iterative closest point (ICP) before computing the distance, isolating reconstruction quality from global scale or translation errors.

\label{sec:rendering_metrics}
We also assess the photometric fidelity of rendered reconstructions. 
We compute three standard image-quality metrics between rendered outputs and ground-truth RGB: PSNR, SSIM, and LPIPS. 
Finally, we report the average optimization runtime on a single NVIDIA RTX~A6000 GPU, using an ARCTIC sequence comprising 300 frames of bimanual hand–object interaction. 
This comparison highlights the relative efficiency of our Gaussian-based formulation compared to prior NeRF- and diffusion-based pipelines.


\begin{figure}[t]
    \centering
    \includegraphics[width=0.85\columnwidth]{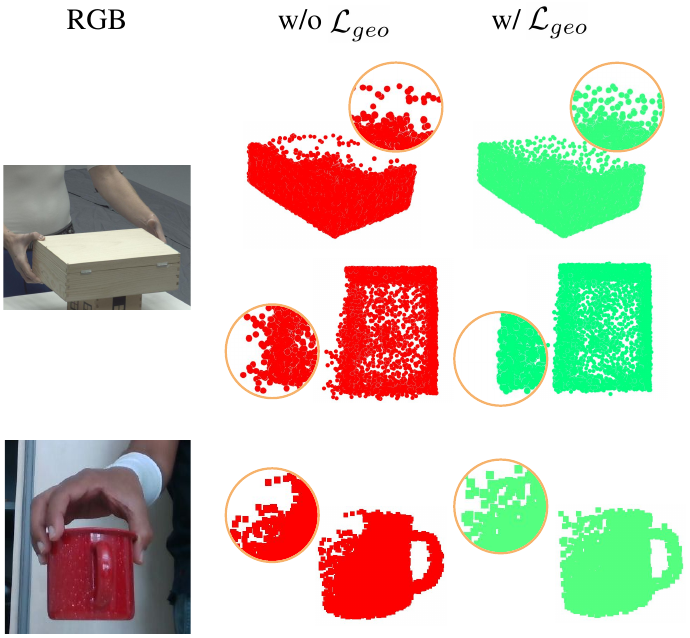}
    \caption{
    Qualitative comparison demonstrating the effect of the geometric loss $\mathcal{L}_{geo}$ on the quality of reconstructed object point clouds derived from Gaussian centers.
    \vspace{-3mm}
    }
    \label{fig:ablation_geo}
\end{figure}



\begin{table}[b]
\vspace{-3mm}
\centering
\caption{Ablation study on the impact of SfM method and geometric consistency loss on object reconstruction and interaction scores for the ARCTIC dataset. $\mathcal{\tau}_{fill}$ is reported in $mm$, $\mathcal{\tau}_{out}$ is in $cm$, and CD metrics are in $cm^2$.
}
\resizebox{\columnwidth}{!}{%
\begin{tabular}{cccccc}
\hline
SfM
& $\mathcal{L}_{geo}$
& $\mathcal{\tau}_{fill}$ 
& $\mathcal{\tau}_{out}$ 
& CD$_h$ $\downarrow$ 
& CD$_{ICP}$ $\downarrow$ \\
\hline
HLoc~\cite{schoenberger2016sfm,sarlin2019coarse,sarlin2020superglue} & \xmark & - & - & - & 8.18 \\
HLoc~\cite{schoenberger2016sfm,sarlin2019coarse,sarlin2020superglue} & \cmark & - & - & - & 4.86 \\
VGGSfM~\cite{wang2024vggsfm}  & \xmark & - & - & 18.32 & 2.78 \\
VGGSfM~\cite{wang2024vggsfm} & \cmark & 1.0 & 5.0 & \textbf{18.21} & 3.66 \\
VGGSfM~\cite{wang2024vggsfm}  & \cmark & 0.1 & 2.0 & 18.42 & 2.33 \\
VGGSfM~\cite{wang2024vggsfm}  & \cmark & 0.01 & 2.0 & 18.37 & 2.83 \\
VGGSfM~\cite{wang2024vggsfm}  & \cmark & 0.01 & 7.0 & 18.4 & \textbf{2.26} \\
\hline
\end{tabular}%
}
\label{tab:ablation}
\end{table}

\begin{table*}[ht!]
\centering
\caption{Comparison of 3D metrics across HOLD, BIGS, and our method. All MPJPE metrics are in $mm$, and all CD metrics are in $cm^2$.}
\resizebox{\textwidth}{!}{%
\begin{tabular}{lcccccccccc}
\hline
Method 
& $\text{MPJPE}_{RA,h}$$\downarrow$
& $\text{MPJPE}_{RA,l}$$\downarrow$
& $\text{MPJPE}_{RA,r}$$\downarrow$
& $\text{CD}_{ICP}$$\downarrow$
& $\text{CD}_r$$\downarrow$
& $\text{CD}_l$$\downarrow$
& $\text{CD}_h$$\downarrow$
& $\text{F}_{10mm}$ (\%)$\uparrow$
& $\text{F}_{5mm}$ (\%)$\uparrow$ \\
\hline
HOLD~\cite{fan2024hold} 
& 25.91 & 27.13 & 24.70 
& \cellcolor{yellow!15}2.07 
& 123.54 & 105.92 & 114.73 
& \cellcolor{yellow!15}63.92 & \cellcolor{yellow!15}37.13 \\

BIGS~\cite{on2025bigs} 
& \cellcolor{yellow!15}24.49 & \cellcolor{green!15}\textbf{24.63} & \cellcolor{yellow!15}24.35 
& \cellcolor{green!15}\textbf{1.36} 
& \cellcolor{yellow!15}31.28 & \cellcolor{yellow!15}46.11 & \cellcolor{yellow!15}38.69 
& \cellcolor{green!15}\textbf{81.78} & \cellcolor{green!15}\textbf{56.41} \\

\textbf{Ours} 
& \cellcolor{green!15}\textbf{24.07} & \cellcolor{yellow!15}25.42 & \cellcolor{green!15}\textbf{22.71} 
& 2.26 
& \cellcolor{green!15}\textbf{13.40} & \cellcolor{green!15}\textbf{23.41} & \cellcolor{green!15}\textbf{18.40} 
& 60.88 & 34.67 \\
\hline
\end{tabular}%
}
\label{tab:sota}
\end{table*}

\begin{table*}[h!]
\centering
\caption{Evaluation of 2D rendering quality on ARCTIC and HO3D datasets, comparing our method with prior works. 
In addition, we report significant runtime improvement.}
\resizebox{0.8\textwidth}{!}{%
\begin{tabular}{l|ccc|ccc|c}
\hline
\multirow{2}{*}{Method} 
& \multicolumn{3}{c|}{\textbf{2D Rendering Quality (ARCTIC)}} 
& \multicolumn{3}{c|}{\textbf{2D Rendering Quality (HO3D)}} 
& \multirow{2}{*}{Runtime$\downarrow$} \\
\cline{2-7}
 & PSNR$\uparrow$ & SSIM$\uparrow$ & LPIPS$\downarrow$
 & PSNR$\uparrow$ & SSIM$\uparrow$ & LPIPS$\downarrow$ &  \\
\hline
HOLD~\cite{fan2024hold} 
& 12.83 & 0.66 & 0.32 
& 16.20 & 0.74 & 0.21 
& 16h \\

BIGS~\cite{on2025bigs} 
& \cellcolor{yellow!15}24.87 & \cellcolor{green!15}\textbf{0.96} & \cellcolor{yellow!15}0.05 
& \cellcolor{green!15}\textbf{24.51} & \cellcolor{green!15}\textbf{0.92} & \cellcolor{yellow!15}0.07 
& \cellcolor{yellow!15}13h \\

\textbf{Ours} 
& \cellcolor{green!15}\textbf{25.93} & \cellcolor{yellow!15}0.88 & \cellcolor{green!15}\textbf{0.02} 
& \cellcolor{yellow!15}21.37 & \cellcolor{yellow!15}0.75 & \cellcolor{green!15}\textbf{0.03} 
& \cellcolor{green!15}\textbf{1h} \\
\hline
\end{tabular}%
}
\label{tab:rendering_table}
\end{table*}

\begin{figure}[t]
    \centering
    \includegraphics[width=0.85\columnwidth]{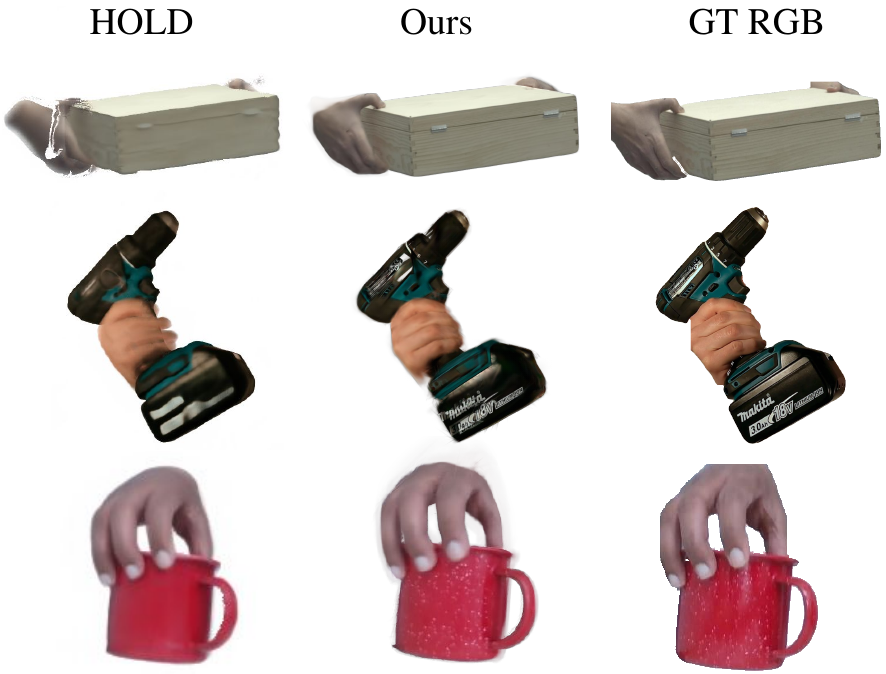}
    \caption{
    \textbf{Qualitative comparison of 2D rendered images:} We compare our results against \emph{HOLD}~\cite{fan2024hold} using representative examples from the evaluated datasets. 
    As shown, \emph{ours} produces higher quality and more consistent 2D renderings, whereas \emph{HOLD} exhibits noise, blur, and degraded visual quality.
    }
    \label{fig:hold_comp}
\end{figure}
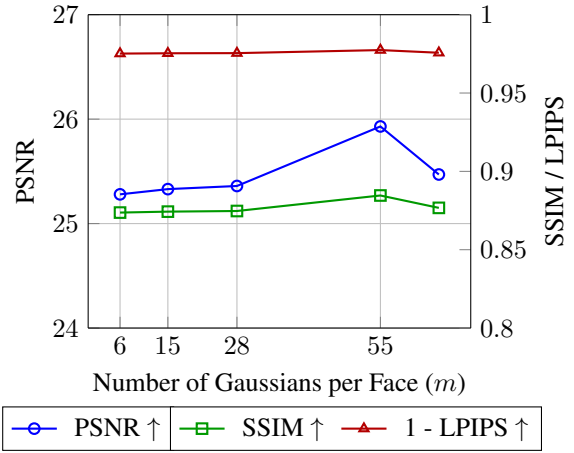
\begin{figure}[t]
\centering
\begin{tikzpicture}

\begin{axis}[
    width=0.8\columnwidth,
    xlabel={Number of Gaussians per Face ($m$)},
    ylabel={PSNR},
    ymin=24, ymax=27,
    xtick={6,15,28,55},
    axis y line*=left,
    ymajorgrids=true,
    xmajorgrids=true,
    legend style={
        at={(0.0,-0.25)},
        anchor=north,
        legend columns=3,
        column sep=0.15cm
    },
]
\addplot[
    thick,
    mark=o,
    color=blue,
] coordinates {
    (6,25.28)
    (15,25.33)
    (28,25.36)
    (55,25.93)
    (66,25.47)
};
\addlegendentry{PSNR $\uparrow$};
\end{axis}

\begin{axis}[
    width=0.8\columnwidth,
    ylabel={SSIM / LPIPS},
    ymin=0.8, ymax=1,
    axis y line*=right,
    axis x line=none,
    legend style={
        at={(0.7,-0.25)},
        anchor=north,
        legend columns=3,
        column sep=0.15cm
    },
]

\addplot[
    thick,
    mark=square,
    color=green!60!black,
] coordinates {
    (6,0.8737)
    (15,0.8743)
    (28,0.8747)
    (55,0.8846)
    (66,0.8767)
};
\addlegendentry{SSIM $\uparrow$};

\addplot[
    thick,
    mark=triangle,
    color=red!70!black,
] coordinates {
    
    (6, 0.97517)
    (15, 0.97537)
    (28, 0.97546)
    (55, 0.97741)
    (66, 0.9757)
};
\addlegendentry{1 - LPIPS $\uparrow$};

\end{axis}

\end{tikzpicture}
\caption{Rendering quality as a function of surface Gaussian density. Higher density improves reconstruction fidelity.
\vspace{-2mm}
}
\label{fig:gaussian_face_density_metrics}
\end{figure}

\subsection{Ablation Study}

\paragraph{Structure-from-Motion (SfM).}
In Table~\ref{tab:ablation}, we evaluate the impact of different SfM methods and the geometric consistency loss $\mathcal{L}_{geo}$ on object reconstruction metrics and interaction metrics. 
Category-agnostic object reconstruction from videos is very sensitive to the object's Structure-from-Motion (SfM) as noted in previous methods~\cite{fan2024hold,on2025bigs}. 
Therefore, we compare two SfM pipelines (i.e., HLoc combined with COLMAP~\cite{sarlin2019coarse, sarlin2020superglue,schoenberger2016sfm} and VGGSfM~\cite{wang2024vggsfm}). 
We observed significant improvements using VGGSfM on the ARCTIC data compared to the HLoc pipeline. 
However, VGGSfM did not show similar improvements on the HO3D dataset sequences~\cite{hampali2020honnotate} and our own recorded data.

\paragraph{$\boldsymbol{\mathcal{L}}_{{geo}}$ and $\boldsymbol{\mathcal{L}}_{{bkg,h}}$.}

The purpose of $\mathcal{L}_{geo}$ is to retrieve occluded regions of the object. 
Fig.~\ref{fig:ablation_geo} illustrates the impact of applying $\mathcal{L}_{geo}$ on the point cloud obtained from Gaussian centers $c_o$. 
It shows that point clouds become clustered around their priors completing missing regions of the object. 
Table~\ref{tab:ablation} compares how $\mathcal{L}_{geo}$ and its hyperparameters ($\tau_{fill}$ and $\tau_{out}$) affect the object reconstruction for the ARCTIC dataset. 
Decreasing $\tau_{out}$ keeps the object's point cloud closer to the prior, restricting the freedom of the gaussians to roam away from the prior where needed. 
Lower $\tau_{fill}$ results in more holes in the point cloud being filled.
Note that we only apply $\mathcal{L}_{geo}$ on sequences where high-quality geometric priors were found for the object. 
Supplementary material also includes more evaluation on the impact of $\mathcal{L}_{geo}$ on a selected set of HO3D sequences. 
On the other side, applying $\mathcal{L}_{bkg,h}$ focuses on retrieving hand-covered object parts. 
Fig.~\ref{fig:ablation_bkg} illustrates how $\mathcal{L}_{bkg,h}$ improved the reconstruction of the object under hand occlusion.

\paragraph{Number of Gaussians per hand ($m$).}
Furthermore, Fig. ~\ref{fig:gaussian_face_density_metrics} shows the importance of the number of Gaussians ($m$) attached to hand mesh faces on the rendering quality of the final hand-object reconstructions. 
We observe that using $55$ Guassians per face and in total $84k$ Gaussians per hand produces the highest rendering quality. 
Because the number of training iterations is fixed ($30$k), adding more Gaussians makes them harder to fit properly. 
The optimization spends more time adjusting the larger set, and the shape becomes more complex, which eventually reduces accuracy.


\subsection{Comparison with state-of-the-art}
Our approach consistently outperforms previous methods on the ARCTIC dataset, achieving lower interaction metrics (CD$_h$, CD$_r$, and CD$_l$) as reported in Table~\ref{tab:sota}.
This improvement is attributed to the grasp-aware contact loss in HO alignment and the novel object reconstruction losses. 
Furthermore, we show the importance of jitter detection discussed in Section~\ref{sec:hamer} on the improvement of hand reconstruction errors (MPJPE), knowing that previous methods also use HaMeR~\cite{pavlakos2024reconstructing} for hand initialization. 
In addition, we observe an improvement in interaction score (CD$_r$) on $3$ sequences and an improvement in CD$_{ICP}$ on $2$ sequences from the HO3D dataset~\cite{hampali2020honnotate} compared to HOLD~\cite{fan2024hold}.

To evaluate the rendering quality of our GS pipeline, we report 2D rendering quality metrics in comparison with previous methods in Table~\ref{tab:rendering_table}. 
We observe an improvement in PSNR and LPIPS compared to BIGS~\cite{on2025bigs} and a significant improvement compared to HOLD in all metrics on the ARCTIC dataset~\cite{fan2023arctic}. 
This can also be observed qualitatively in Fig.~\ref{fig:hold_comp}, where we show different rendering examples in comparison to HOLD. 
All this is obtained while achieving 13× improvement in runtime as reported in Table~\ref{tab:rendering_table}. 
More qualitative results are shown in Figs.~\ref{fig:teaser} and ~\ref{fig:qualitative}.

\vspace{-2mm}
\section{Conclusion}


In this paper, we introduced GHOST, a fast and category-agnostic framework for reconstructing realistic hand–object interactions from monocular RGB videos using 2D Gaussian Splatting. 
By combining geometric-prior retrieval, grasp-aware alignment, and hand-aware background reasoning, GHOST produces complete object surfaces, physically consistent hand–object contact, and animatable hand avatars. 
GHOST sets a new state-of-the-art baseline in 3D interaction reconstruction accuracy and 2D rendering quality on the ARCTIC Bi-CAIR benchmark, while running over 13× faster than existing category-agnostic baselines.
Beyond quantitative gains, GHOST also produces animatable hand avatars and photorealistic novel-view renderings.
In future work, we will extend GHOST to deformable and articulated objects, explore direct integration of geometric priors within the SfM pipeline, and investigate real-time inference from stereo or RGB-D streams. 
These extensions can enable practical deployment in teleoperation, interactive AR/VR systems, robotics manipulation, and in-the-wild motion capture settings where speed and physical plausibility are critical.
\textbf{Acknowledgments:}
This work was partially funded by the European Union's Horizon Europe programme through the SHARESPACE project (grant no. 101092889) and the LUMINOUS project (grant no. 101135724).



\clearpage
{
    \small
    \bibliographystyle{ieeenat_fullname}
    \bibliography{GHOST/references}
}

\clearpage
\setcounter{page}{1}
\maketitlesupplementary
\appendix

\begin{figure}[]
    \centering
    \includegraphics[width=0.95\columnwidth]{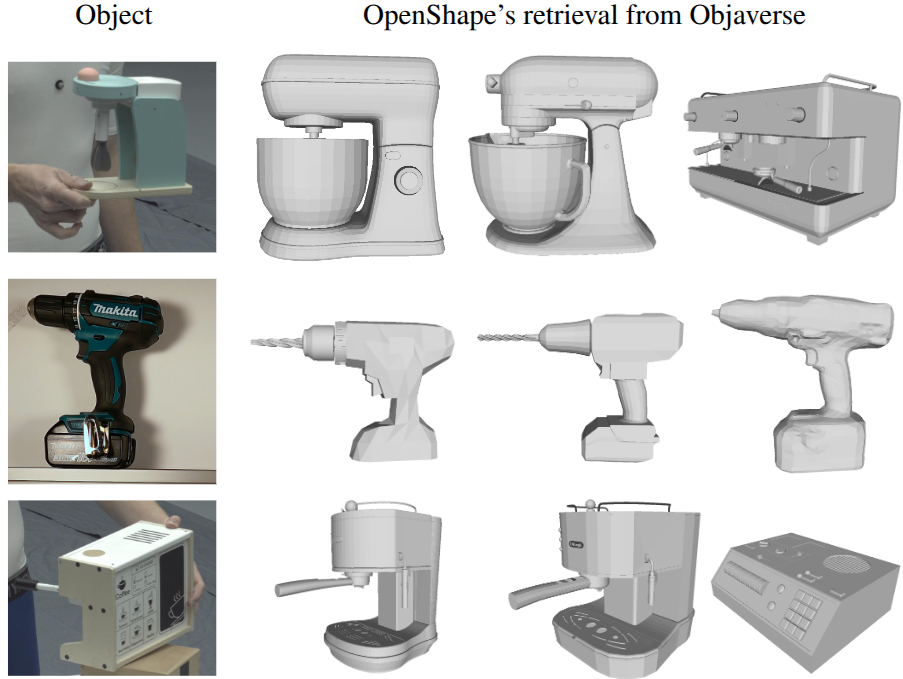}
    \caption{OpenShape~\cite{liu2023openshape} retrieves 3D models from Objaverse~\cite{deitke2023objaverse,deitke2023objaversexl}, however, the retrieved 3D models do not always match the geometry of the desired object. Therefore, the final geometric prior $\mathcal{O}$ can be suboptimal.}
    \label{fig:retrieval_examples}
\end{figure}

\begin{figure}[!t]
    \centering
    \includegraphics[width=\linewidth]{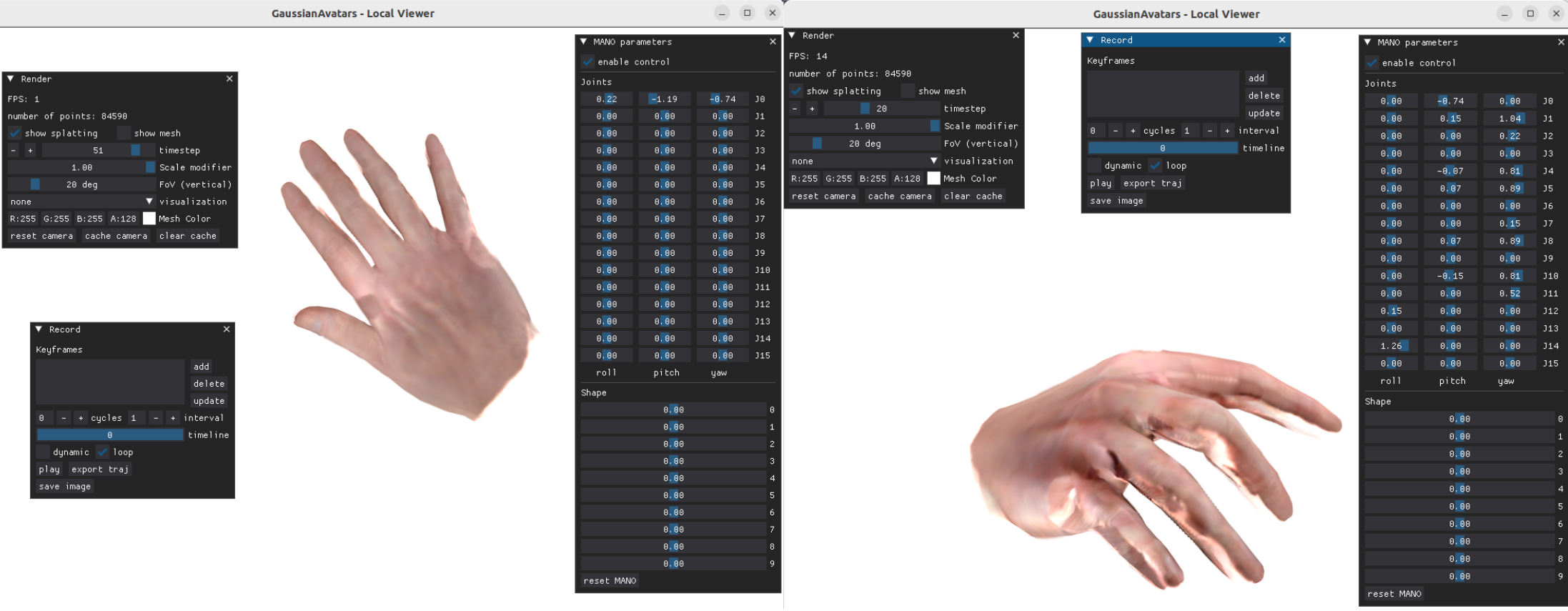}
    \caption{
    We also provide an interactive 3D viewer for Gaussian hand avatars. The interface visualizes and controls MANO-based~\cite{romero2017mano} hand reconstructions, allowing users to adjust both pose and shape parameters in real time. Imported motion sequences can be played to animate the Gaussian hand avatar.
    }
    \label{fig:viewer}
\end{figure}

In our supplementary material, we give more illustrations of different steps in our preprocessing pipeline in Figs.~\ref{fig:illustration} and ~\ref{fig:retrieval_examples}. 
Furthermore, we compare different design choices and their qualitative gains in Fig.~\ref{fig:ablation_supp}. 
Fig.~\ref{fig:ho3d_barplot} shows quantitative evaluation of applying $\mathcal{L}_{geo}$ on 5 sequences from the HO3D dataset~\cite{hampali2020honnotate}. 
We observe an improvement in the interaction distance relative to the right hand root (CD$_r$) on $3$ sequences. 
Fig~\ref{fig:limitations} shows limitations of $\mathcal{L}_{bkg,h}$ and $\mathcal{L}_{geo}$. 
In addition, Appendix~\ref{sec:jitter} and~\ref{sec:experimental} discuss different hyperparameters in our approach. 
Finally, we show examples of our animatable hand avatar visualization inherited from GaussianAvatars~\cite{qian2024gaussianavatars} viewer in Fig.~\ref{fig:viewer}.

\section{Postprocessing HaMeR Predictions}
\label{sec:jitter}

For image $I_t$ with area $A_{img}$, RTMPose~\cite{jiang2023rtmpose,jiang2023rtmlib} provides hand keypoints for left and right hands with confidence for each keypoint producing right and left hand bounding boxes ($B_t^r$, $B_t^l$) with areas ($A_t^r$, $A_t^l$) and averaged keypoint confidence ($c_t^l$ and $c_t^r$). 
Using the hand bounding boxes, HaMeR~\cite{pavlakos2024reconstructing} generates initial hand reconstructions as stated in Section ~\ref{sec:hamer}. 
For timestep $t$, our algorithm decides based on the combined rejection rule in Eq.~\ref{eq:reject} for each hand $h \in \{r,l\}$ if the predictions of the frames should be discarded and interpolated or not. 
The individual conditions are defined as follows:

\begin{figure}[t]
    \centering

    \begin{minipage}[t]{0.5\columnwidth}
        \includegraphics[width=\textwidth]{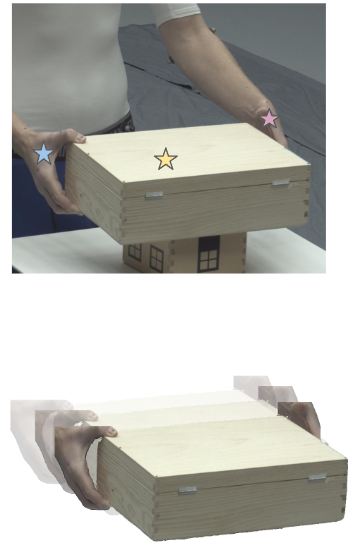}
        \caption*{(a)}
    \end{minipage}
    \hspace{0.01\columnwidth}
    \begin{tikzpicture}
        \draw[dashed, gray!60, line width=0.5pt] (0,0) -- (0,6.5);
    \end{tikzpicture}
    \hspace{0.01\columnwidth}
    \begin{minipage}[t]{0.385\columnwidth}
        \includegraphics[width=\textwidth]{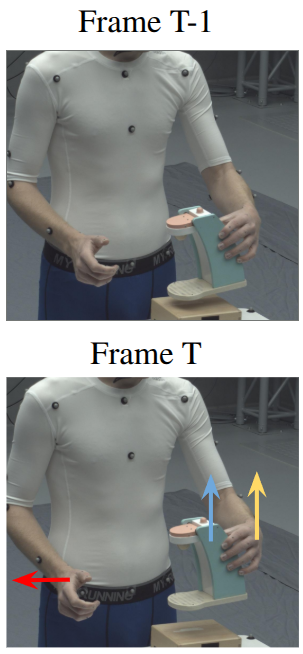}
        \caption*{(b)}
    \end{minipage}

    \caption{
        (a) SAM2~\cite{ravi2024sam2} is initialized with 3 seed pixels to segment and track the hands ($\mathcal{M}_t^h$) and the object ($\mathcal{M}_t^o$) in the scene.  
        (b) During the grasping detection, the object's motion vector $\hat{\mathcal{T}}^{o}_{xy}$ (blue arrow) is compared with left hand's motion vector $\hat{\mathcal{T}}^{l}_{xy}$  (orange arrow) and right hand's motion vector $\hat{\mathcal{T}}^{r}_{xy}$ (red arrow). 
        The example shows a left hand grasp based on the similarity between motion vectors.
    }
    \label{fig:illustration}
\end{figure}

\begin{figure*}[t]
    \centering
    \begin{subfigure}[t]{0.47\textwidth}
        \centering
        \includegraphics[width=\linewidth]{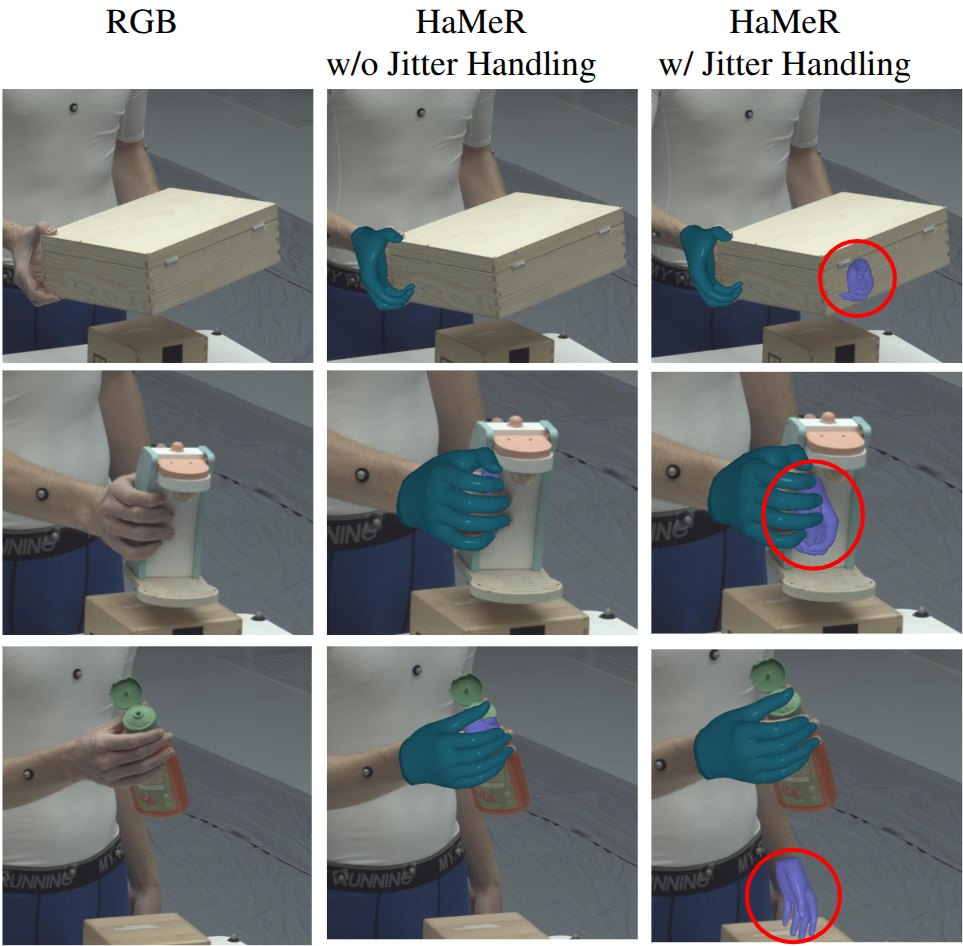}
        \caption{}
        \label{fig:ablation_hamer}
    \end{subfigure}
    \hspace{0.01\columnwidth}
    \begin{tikzpicture}
        \draw[dashed, gray!60, line width=0.5pt] (0,0) -- (0,7.5);
    \end{tikzpicture}
    \hspace{0.01\columnwidth}
    \begin{subfigure}[t]{0.465\textwidth}
        \centering
        \includegraphics[width=\linewidth]{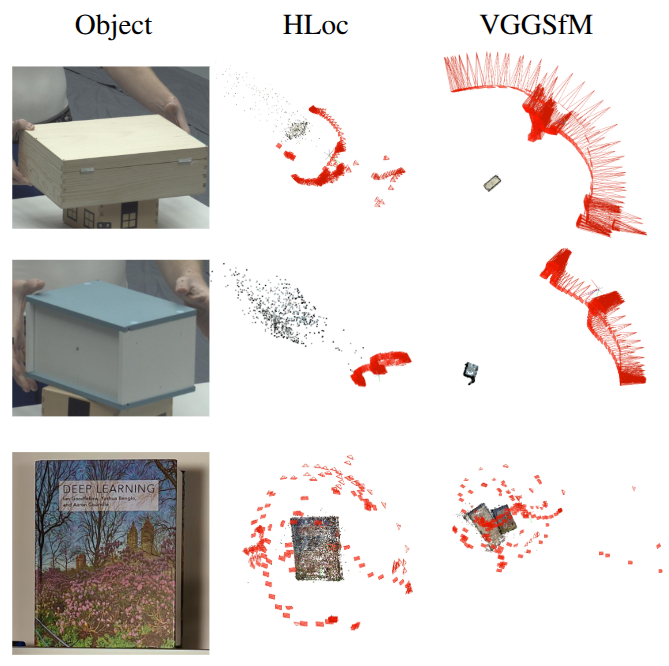}
        \caption{}
        \label{fig:ablation_sfm}
    \end{subfigure}

    \caption{
    a) Initial hand reconstructions $\mathcal{V}_t^h$ obtained from HaMeR~\cite{pavlakos2024reconstructing} suffer from Jitter under occlusion. 
    Detecting jitter based on temporal cues, detection confidence, and interpolation results in improving initial hand meshes $\mathcal{V}_t^h$. 
    b) Structure-from-Motion (SfM) has a large impact on subsequent steps. 
    VGGSfM~\cite{wang2024vggsfm} improved SfM when applied to the arctic data compared to the HLoc+COLMAP~\cite{schoenberger2016sfm,sarlin2019coarse} pipeline. 
    However, VGGSfM is sensitive to hyperparameter selection and does not always show similar improvements as seen in the last row.
    }
    \label{fig:ablation_supp}
\end{figure*}

\pgfplotsset{compat=1.18}

\begin{figure}[h!]
\centering
\begin{tikzpicture}
\begin{axis}[
    ybar,
    bar width=6pt,
    width=0.9\columnwidth,
    height=5.4cm,
    ymin=0,
    ylabel={Chamfer Distance (CD$_r$)}, 
    symbolic x coords={mug,mustard,bleach,crackers,meat},
    xtick=data,
    xticklabel style={align=center, yshift=0pt},
    xticklabels={
        \includegraphics[width=0.3in]{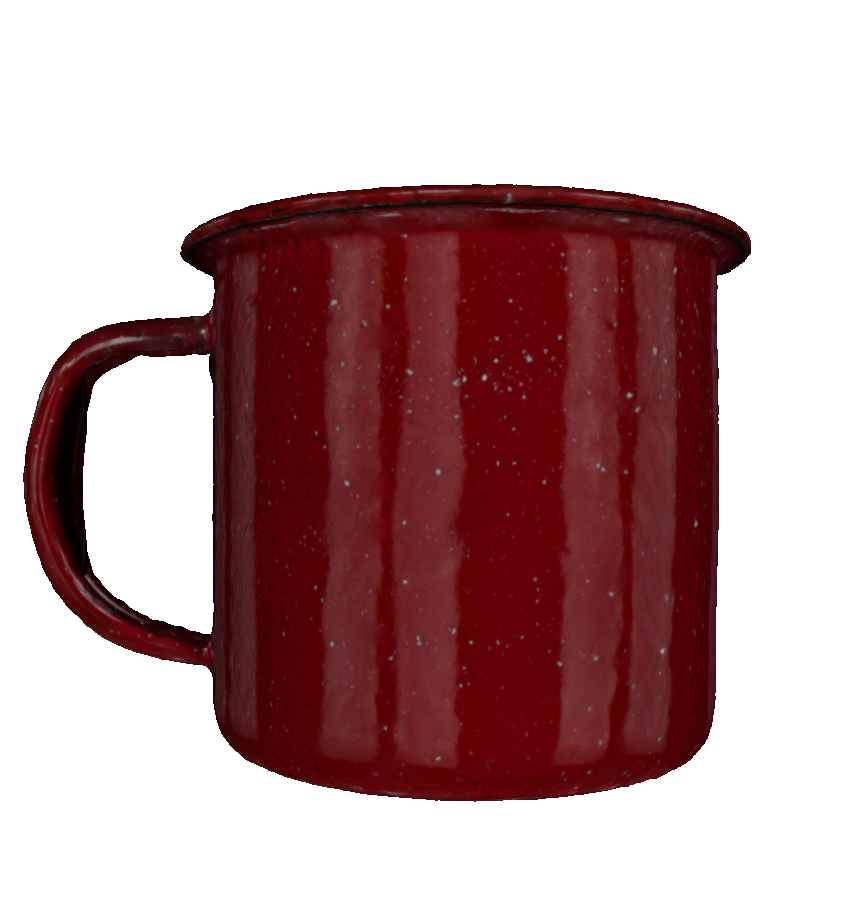},
        \includegraphics[width=0.3in]{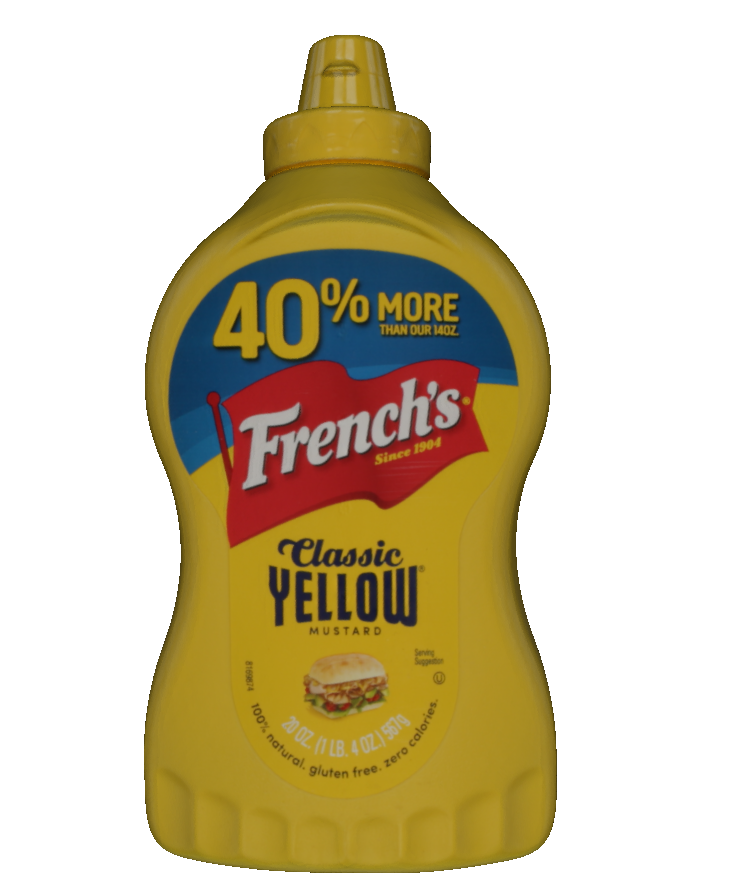},
        \includegraphics[width=0.3in]{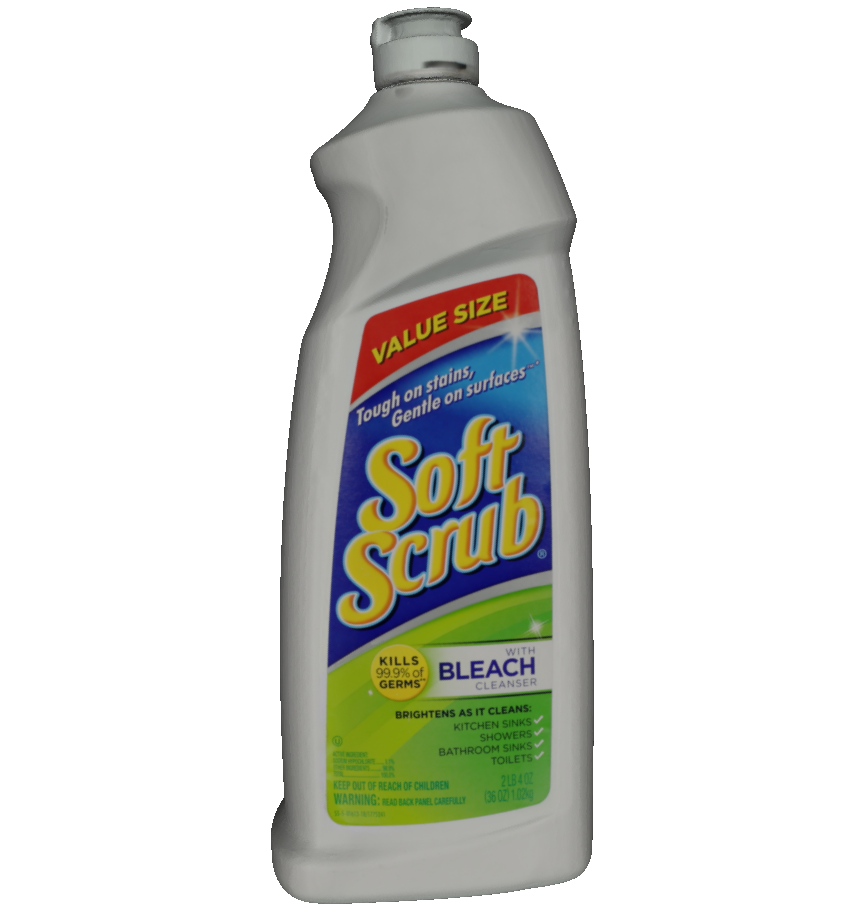},
        \includegraphics[width=0.3in]{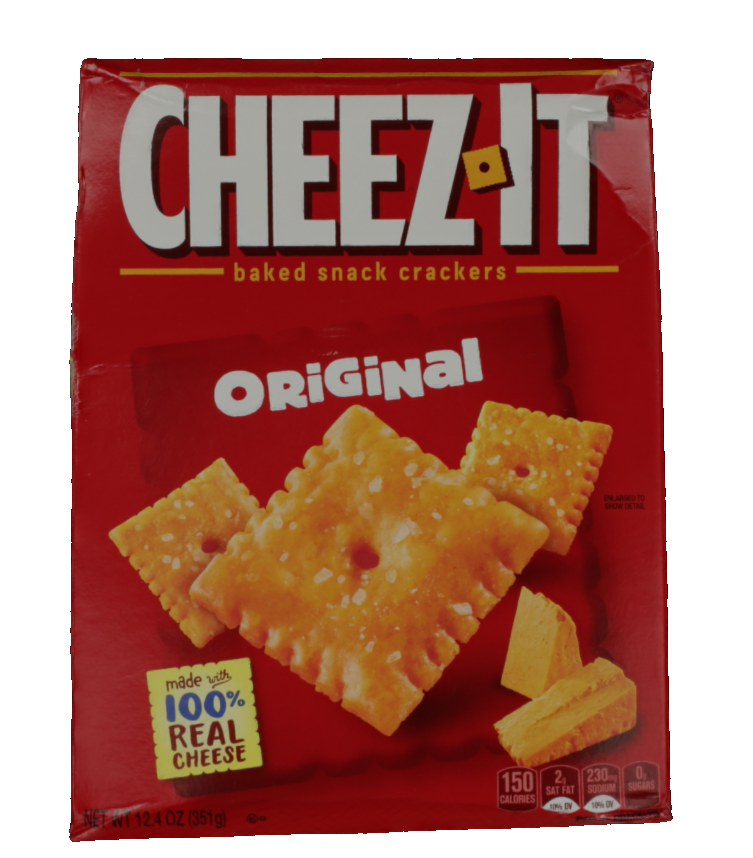},
        \includegraphics[width=0.3in]{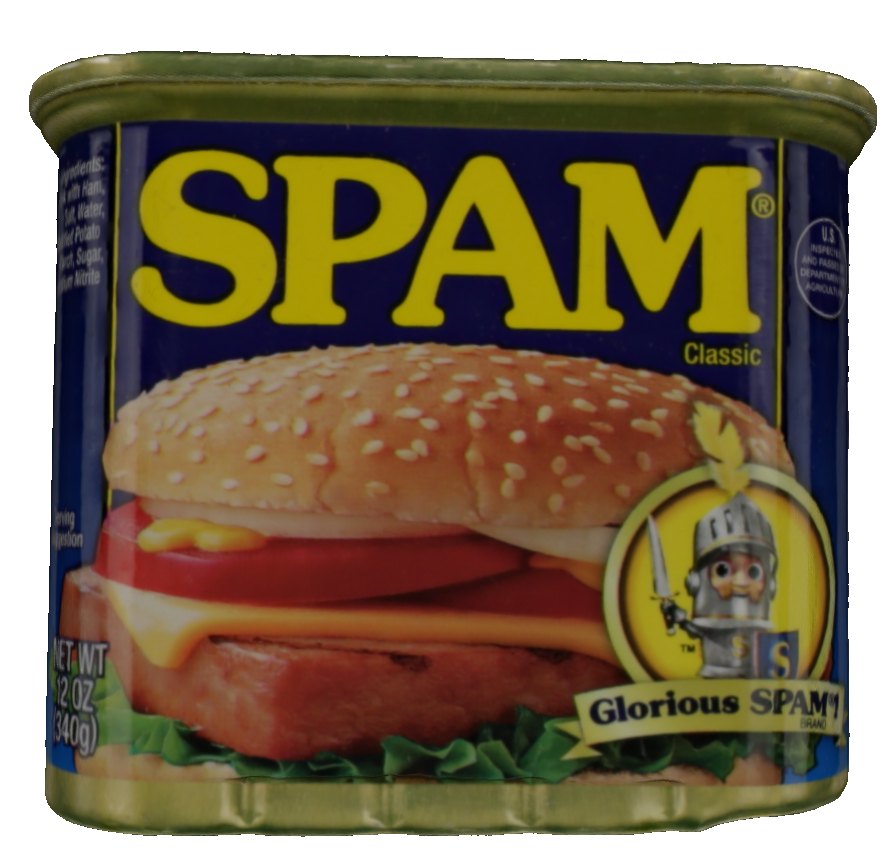}
    },
    ymajorgrids=true,
    grid style=dashed,
    enlarge x limits=0.2,
    legend style={
        at={(0.5,-0.38)},
        anchor=north,
        legend columns=-1,
        font=\small
    },
    clip=false,
]



\addplot+[fill={rgb,255:red,216; green,175; blue,74}, draw=black] coordinates {
    (mug,1.58)
    (mustard,4.53)
    (bleach,5.80)
    (crackers,16.69)
    (meat,18.57)
};
\addplot+[fill={rgb,255:red,102; green,178; blue,178}, draw=black] coordinates {
    (mug,1.47)
    (mustard,4.58)
    (bleach,5.88)
    (crackers,15.98)
    (meat,17.79)
};


\legend{
Ours (w/o $\mathcal{L}_{geo}$),
Ours (w/ $\mathcal{L}_{geo}$)
}

\end{axis}
\end{tikzpicture}
\caption{The influence of $\mathcal{L}_{geo}$ (CD$_r$, lower is better) across five HO3D sequences using distinct YCB objects.
}
\label{fig:ho3d_barplot}
\end{figure}



\begin{figure}[t]
    \centering
    \begin{subfigure}[t]{0.35\columnwidth}
        \centering
        \includegraphics[width=\linewidth]{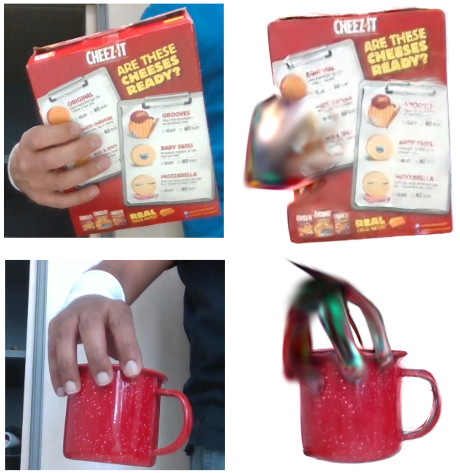}
        \caption{}
        \label{fig:ablation_bkg_supp}
    \end{subfigure}
    \hspace{0.01\columnwidth}
    \begin{tikzpicture}
        \draw[dashed, gray!60, line width=0.5pt] (0,0) -- (0,3);
    \end{tikzpicture}
    \hspace{0.01\columnwidth}
    \begin{subfigure}[t]{0.55\columnwidth}
        \centering
        \includegraphics[width=\linewidth]{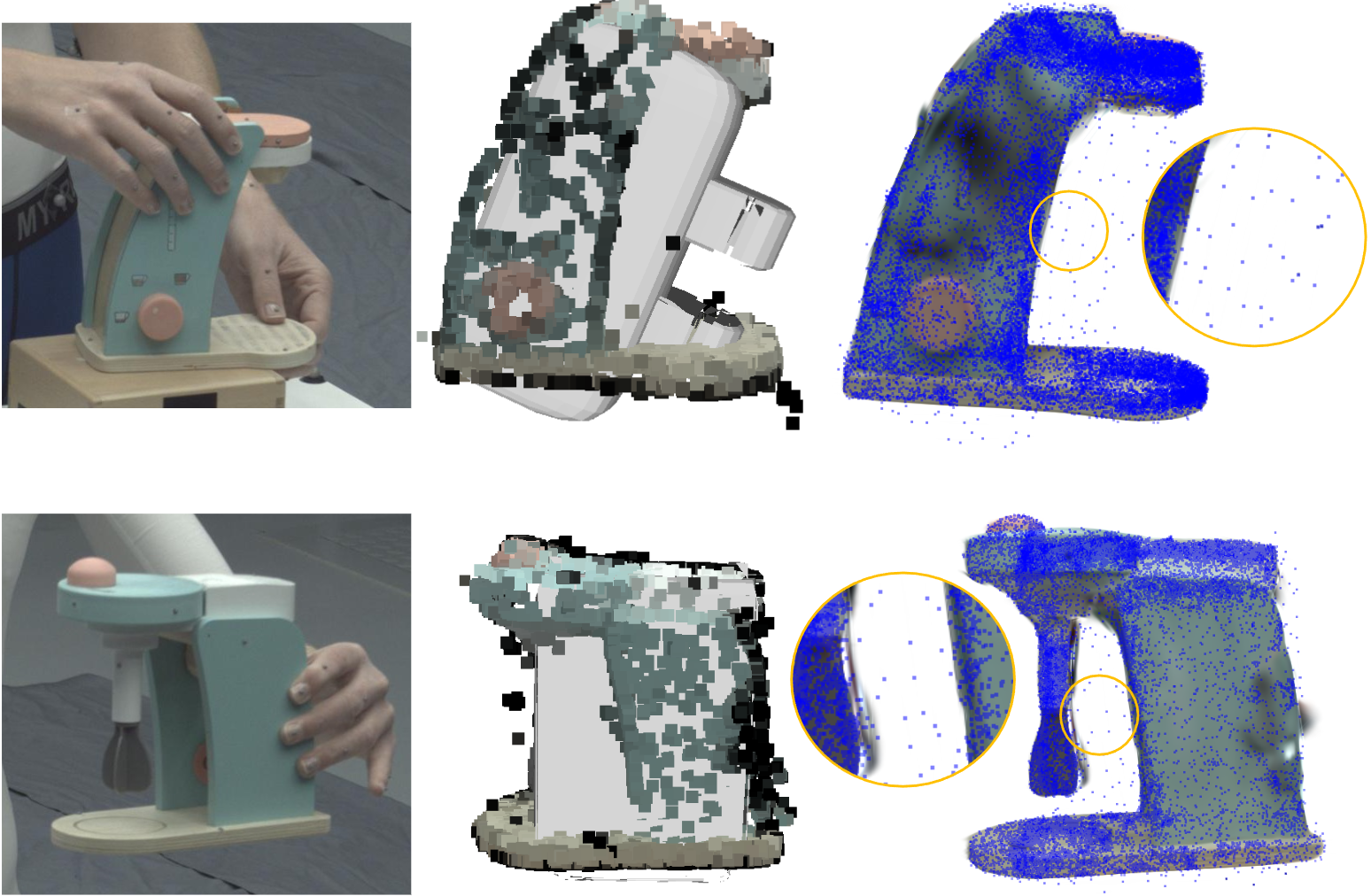}
        \caption{}
        \label{fig:l_geo_supp}
    \end{subfigure}

    \caption{
    a) Applying $\mathcal{L}_{bkg,h}$ for object reconstruction fails when the hand never changes its contact point with the object. 
    In that case, unwanted gaussians spawn in the hand region. 
    b) In some cases, the retrieved geometric prior $\mathcal{O}$ does not align perfectly with the initial object's point cloud $\mathcal{P}_{sfm}$ (Middle column). 
    The results of applying $\mathcal{L}_{geo}$ in this case will result in moving gaussian centers $c_o$ towards unwanted regions (see blue point cloud).
    }
    \label{fig:limitations}
\end{figure}

\noindent
\begin{enumerate}
    \item 
{Pose jitter condition:}
\begin{equation}
\mathcal{C}_{\text{p}} =
\left(
\|\boldsymbol{\theta}_t - \boldsymbol{\theta}_{t-1}\|_2 > \tau_{\text{p}}
\ \wedge\
\|\boldsymbol{\theta}_t - \boldsymbol{\theta}_{t+1}\|_2 > \tau_{\text{p}}
\right),
\end{equation}

\item {Orientation jitter condition:}
\begin{equation}
\mathcal{C}_{{o}} =
\left(
\|\mathbf{R}_t^h - \mathbf{R}_{t-1}^h\|_2 > \tau_{{o}}
\ \wedge\
\|\mathbf{R}_t^h - \mathbf{R}_{t+1}^h\|_2 > \tau_{{o}}
\right),
\end{equation}

\item {Translation jitter condition (x–y plane):}
\begin{equation}
\mathcal{C}_{{t}} =
\left(
\|\mathbf{T}^{h}_{t} - \mathbf{T}^{h}_{t-1}\|_2 > \tau_{{t}}
\ \wedge\
\|\mathbf{T}^{h}_{t} - \mathbf{T}^{h}_{t+1}\|_2 > \tau_{{t}}
\right),
\end{equation}

\item {Shape deviation condition:}
\begin{equation}
\mathcal{C}_{{s}} =
\frac{
\big| \boldsymbol{\beta}_{t} - \operatorname{median}(\boldsymbol{\beta}) \big|
}{
\operatorname{std}(\boldsymbol{\beta}) + \varepsilon
}
> \tau_{{shape}}.
\end{equation}

\item {Confidence threshold:}
\begin{equation}
\mathcal{C}_{{c}} =
{c}_t^h < \tau_{{conf}},
\end{equation}

\item {Bounding-box area constraint:}
\begin{align}
\mathcal{C}_{{a}} =
\left(
A_t^h < A_{\min}
\ \vee\
A_t^h > A_{\max}
\right),
\end{align}

\item {Bounding-box overlap (IoU) constraint:}
\begin{equation}
\mathcal{C}_{{iou}} =
IoU(B_t^r,B_t^l),
\end{equation}
\end{enumerate}

\noindent where the thresholds are empirically chosen as: 
$\tau_{\text{p}} = 1.0$, $\tau_{\text{o}} = 1.0$, $\tau_{\text{t}} = 2.0$, $\tau_{\text{s}} = 4.0$, $\tau_{\text{c}} = 0.3$, $\tau_{\text{iou}} = 0.3$, $A_{\min} = 0.006\,A_{{img}}$, and $A_{\max} = 0.2\,A_{{img}}$.

\noindent\textbf{Hand rejection rule:}
\begin{equation}
\label{eq:reject}
\mathcal{F}_{\text{reject}} =
\mathcal{C}_{\text{p}}
\ \vee\
\mathcal{C}_{\text{o}}
\ \vee\
\mathcal{C}_{\text{t}}
\ \vee\
\mathcal{C}_{\text{s}}
\ \vee\
\mathcal{C}_{\text{c}}
\ \vee\
\mathcal{C}_{\text{a}}
\ \vee\
\mathcal{C}_{\text{iou}}.
\end{equation}

Fig~\ref{fig:ablation_hamer} shows the importance of this rejection rule on hand meshes $\mathcal{V}_t^h$.

\section{Experimental details}
\label{sec:experimental}

\begin{enumerate}
    \item Prior Alignment Parameters. Optimizer: AdamW~\cite{loshchilov2018decoupled}, LR: $10^{-2}$, Betas: $0.9, 0.99$, eps:$10^{-8}$, Iterations: $1500$.
\item HO Alignment.
Optimizer: Adam~\cite{adam2014method}, LR: $0.05$, Iterations: $500$.
\item Object Gaussian Splatting Optimization. Optimizer: Adam~\cite{adam2014method}. Iterations: $30000$.
\item Hand-Object Gaussian Splatting Optimization. Optimizer: Adam~\cite{adam2014method}. Iterations: $30000$.
More details on the Gaussian Splatting hyperparameters are available in the code.
\end{enumerate}

\end{document}